%% file: main.tex
\documentclass[runningheads]{llncs}
\usepackage[T1]{fontenc}
\usepackage{graphicx}

\usepackage[hidelinks]{hyperref}
\usepackage{booktabs}
\usepackage{algorithm}
\usepackage{algorithmic}
\usepackage{mathtools}
\usepackage{amssymb}
\usepackage{tabularx}
\usepackage{multirow}

\usepackage{caption}
\usepackage{subcaption}
\usepackage{comment}
\usepackage{graphicx}
\usepackage{csquotes}
\usepackage{enumitem}

\usepackage{color}

\begin{document}
\title{Revealing Similar Semantics Inside CNNs:\\An Interpretable Concept-based Comparison of Feature Spaces}
\titlerunning{Interpretable Concept-based Comparison of Feature Spaces}
%

\author{
Georgii Mikriukov\inst{1,2}\orcidID{0000-0002-2494-6285}
\and Gesina Schwalbe\inst{1}\orcidID{0000-0003-2690-2478}
\and Christian Hellert\inst{1}\orcidID{0000-0002-5781-6575}
\and Korinna Bade\inst{2}\orcidID{0000-0001-9139-8947} 
}

\authorrunning{G. Mikriukov et al.}

\institute{
Continental AG, Germany\\
\email{\{firstname.lastname\}@continental-corporation.com}\\
\and
Hochschule Anhalt, Germany\\
\email{\{firstname.lastname\}@hs-anhalt.de}
}

\maketitle

\input{sections/00_abstract.tex}

\input{sections/10_intro.tex}

\input{sections/20_related.tex}

\input{sections/30_background.tex}

\input{sections/40_method.tex}

\input{sections/50_setup.tex}

\input{sections/60_experiments.tex}

\input{sections/70_conclusion.tex}

\input{sections/80_acknowledgments.tex}

\newpage
\bibliographystyle{splncs04}
\bibliography{ref}
\end{document}

%% file: sections/00_abstract.tex
\begin{abstract}

Safety-critical applications require transparency in artificial intelligence (AI) components, but widely used convolutional neural networks (CNNs) widely used for perception tasks lack inherent interpretability.
Hence, insights into what CNNs have learned are primarily based on performance metrics, because these allow, e.g., for cross-architecture CNN comparison. However, these neglect how knowledge is stored inside.
To tackle this yet unsolved problem, our work proposes two methods for estimating the layer-wise similarity between semantic information inside CNN latent spaces. These allow insights into both the flow and likeness of semantic information within CNN layers, and into the degree of their similarity between different network architectures.
As a basis, we use two renowned explainable artificial intelligence (XAI) techniques,
which are used to obtain concept activation vectors, i.e., global vector representations in the latent space. These are compared with respect to their activation on test inputs.
When applied to three diverse object detectors and two datasets, our methods reveal that (1) similar semantic concepts are learned \emph{regardless of the CNN architecture}, and (2) similar concepts emerge in similar \emph{relative} layer depth, independent of the total number of layers.
Finally, our approach poses a promising step towards semantic model comparability
and comprehension of how different CNNs process semantic information.

\keywords{Explainable Artificial Intelligence \and Network Comparison \and Feature Space Comparison \and Semantic Concept.}

\end{abstract}


%% file: sections/10_intro.tex
\section{Introduction}
\label{sec:intro}

The emerging use of artificial intelligence~(AI) and especially CNNs in safety-critical applications such as automated driving and medicine, has made the interpretability and transparency~\cite{arrieta2020explainable,mittelstadt2019explaining} of these models increasingly essential, not least because industrial and legal standards 
demand sufficient evidence of developed AI modules for safe and ethical use\cite{iso26262,goodman2017european}.
Therefore, it is crucial to develop methods that reveal the model semantics, i.e., what was learned where inside, in particular in relation to other models.
Such model comparability at knowledge level can enhance the general understanding of model knowledge encoding, the influence of architectures, and possibly also datasets. Some potential future applications are retrieval of dataset bias, and informed model selection and architecture modification.

One popular method of knowledge representation assessment within the field of XAI is analysis of semantic concepts, where concepts correspond to real-world objects or notions~\cite{bau2017network,mittelstadt2019explaining,nguyen_understanding_2019}. These concepts are associated with vectors in the CNN feature space, the so-called concept activation vectors (CAV)~\cite{kim2018interpretability,zhang2021invertible}. By examining the CAVs and their responses to model inputs, experts can
gain valuable insights into model operation.

This research proposes two architecture-agnostic strategies for estimating the similarity of feature spaces and semantic concepts in CNNs. These allow to answer for any two CNN layers how similar they are regarding their learned concepts (unsupervised strategy) and regarding any given set of user-defined concepts (supervised strategy). To achieve this, we use the concept analysis methods TCAV~\cite{kim2018interpretability} (supervised) and ICE~\cite{zhang2021invertible} (unsupervised) as the basis. Both generate CAVs for concept-related samples during training. The response of these CAVs to test data is then measured to determine the feature space similarity with respect to the given concepts. The contributions and findings of this work are the following:

\begin{itemize}
    \item We conduct a \textbf{concept-based comparison of feature spaces} and show how the same semantic information is processed differently across various CNN backbones;
    \item For the comparison we propose \textbf{an unsupervised and a supervised layer-wise approach to compare the semantic information} encoded in CNNs, which are shown to yield intuitive and interpretable results regarding CNN knowledge inspection;
    \item The main findings of our concept-based comparison of feature spaces are: \textbf{same semantic concepts are learned across different CNN architectures} and can be extracted from proper layers, representations of \textbf{concepts are located at the same relative depth of the backbone} in feature spaces of different networks.
\end{itemize}

%% file: sections/20_related.tex
\section{Related Work}
\label{sec:related}

\medskip\noindent\textbf{Explainable AI.}
The field of XAI encompasses interpretability techniques~\cite{schwalbe2021comprehensive} to explain the predictions of machine learning functions like neural networks (NNs) to a human.
%
While ante-hoc approaches using models that are interpretable, e.g., produce human-understandable concept outputs~\cite{koh2020concept,losch_interpretability_2019,chen2019looks}, are  preferable~\cite{rudin_stop_2019}, we here concentrate on already trained CNNs.
%
For post-hoc explainability, one can distill approximate interpretable surrogate models. Concept-based examples are 
flow-graphs~\cite{hohman_summit_2020}, 
layer-wise concept hierarchies~\cite{%
wang2022hint,
wang_chain_2020,
zhang_interpreting_2018
},
decision trees~\cite{%
wan_nbdt_2020,
chyung_extracting_2019},
and rule sets~\cite{rabold_expressive_2020,rabold_explaining_2018}.
However,
the limited fidelity to the original CNN renders
them unsuitable for quantitative CNN comparison.
%
Other post-hoc methods concentrate on explaining the behavior for single samples. Such can be applied in an approximate model-agnostic manner~\cite{rabold_explaining_2018,ribeiro2016model} or model-specific based on the model internal processing, like prominent saliency methods~\cite{zhou2016learning,
selvaraju2017grad,
bach2015pixel,
achtibat2022towards
}.
Such local approaches, even if aggregated to global information like in \cite{%
lapuschkin_unmasking_2019
}, only give limited insights into concepts represented in the CNN internals.
Instead, this work relies on concept analysis~\cite{schwalbe_concept_2022}, i.e., XAI methods that allow direct insights into the human-understandable concepts learned by a CNN.

\medskip\noindent\textbf{Concept Analysis.}
Early techniques associate single CNN units with concepts~\cite{bau2017network,nguyen_understanding_2019},
disregarding the distributed nature of CNN representations.
Supervised linear methods like state-of-the-art TCAV~\cite{kim2018interpretability} associate concepts to latent space vectors.
Further extensions to use-cases like concept
regression~\cite{graziani_concept_2020} and localization~\cite{lucieri_explaining_2020} also stuck to this principle. There are also non-linear alternatives like clustering~\cite{%
gu_semantics_2019,
kazhdan_now_2020
} or NNs~\cite{esser_disentangling_2020}, which, however, pose additional requirements to the labels.
Unsupervised approaches require no concept labels at all, like ICE~\cite{zhang2021invertible} that applies matrix factorization to the latent space. Alternatives relying on intelligent choice of concept candidate patches~\cite{%
ghorbani2019towards,
ge2021peek
} lead to less interpretable results~\cite{zhang2021invertible}.

\textbf{Network Comparison.}
Existing neural network comparison methods foremostly utilize performance or error-estimation metrics, and qualitative manual observation based on visual analytics or XAI. 
Examples for object detection model analysis are the TIDE~\cite{bolya2020tide} metrics and visualizations toolbox, and the framework by Miller~et\,al.~\cite{miller2022s} to analyze models' ability to handle false negative occurrences.
More knowledge-based approaches measure the compliance with constraints like object relations~\cite{giunchiglia_roadr_2022,schwalbe_enabling_2022} or temporal consistency~\cite{varghese_unsupervised_2021}.


%% file: sections/30_background.tex
\section{Background}
\label{sec:background}

In contrast to the mentioned methods, our approach involves comparing feature spaces, i.e.,  knowledge encoded in CNNs, through semantic concepts and their responses to various inputs. A (visual) semantic concept refers to a feature of an image that can be expressed in natural language (e.g., \enquote{head} or \enquote{green}) \cite{bau2017network,fong2018net2vec}. 
Concepts can be associated with a numeric vector in the latent space, known as the concept vector \cite{kim2018interpretability,fong2018net2vec}.
The approaches for this used in this paper are shortly recapitulated in the following.

\textbf{TCAV.}
TCAV~\cite{kim2018interpretability} is a supervised concept analysis method that utilizes Concept Activation Vectors (CAVs) to represent concepts in the latent space of a NN. Parameters of CAVs correspond to those of a binary linear classifier that separates the feature space of a given layer in a concept-versus-rest manner. The classifier is trained using the activations of concept-related and unrelated samples. Geometrically, a CAV is the normal vector to the separation hyperplane and indicates the direction of the concept in the latent space. The similitude between a sample and concepts is defined by cosine similarity. This feature of CAVs can be employed for ranking of input samples by concept-relevance.

\textbf{ICE.}
The unsupervised ICE~\cite{zhang2021invertible} approach employs Non-Negative Matrix Factorization (NMF) to mine a pre-defined small number of Non-negative CAVs (NCAVs) in the latent space. These NCAVs correspond to the most frequent patterns of activation in convolutional filters caused by the training samples. NCAVs are then utilized to map input sample activations of dimensionality $C \times H \times W$ to $C^{\prime} \times H \times W$ dimensional concept activations, where $C$, $H$, $W$, and $C^{\prime}$ represent the \textit{channel}, \textit{height}, \textit{width}, and \textit{concept} dimensions, respectively. Each of the $C^{\prime}$ concept activations of size $1 \times H \times W$ is normalized, interpolated to the original input size, and employed as a saliency map to highlight the concept-related regions. The examples of such binarized masks are presented in 
Fig.~\ref{fig:ice-concepts-celeba-and-mscoco}.

%% file: sections/40_method.tex
\section{Semantic Comparability Methods}
\label{sec:method}

To address the gap in literature on comparison of model semantics, we introduce supervised and unsupervised approaches that use concept representations to compare feature spaces of CNN backbones. These methods rely on relative semantic similarity ranking of samples and overlap estimation of concept saliency maps. Section~\ref{sec:method-unsupervised-sim} and Section~\ref{sec:method-supervised-sim} provide details on the unsupervised and supervised comparison approaches, respectively.

\input{graphics/main_diagram_unsupervised}

\subsection{Unsupervised Concept Similarity}
\label{sec:method-unsupervised-sim}

The proposed unsupervised approach addresses two key questions: \textit{\enquote{Are there similar concepts in feature spaces of different layers?}} and \textit{\enquote{How similar are they?}}. We utilize ICE~\cite{zhang2021invertible} to identify and extract the most prominent activation patterns, which are represented by NCAVs, in the feature spaces of different layers. Then, we measure the overlap between binarized concept saliency maps on test data to compare the similarity of extracted concepts in selected layers. Although we use ICE in our work, the general approach is not limited to this specific method, and shall only showcase the usage of saliency methods.  

Figure~\ref{fig:method-unsupervised}a depicts the process of layer-wise unsupervised knowledge comparison in two trained \textit{Tested Networks}, which may have different architectures. \textit{L1} and \textit{L2} are indices of analyzed layers. The first step involves using the activations of training samples (\textit{Train Acts}) obtained from the \textit{Train Images} to automatically extract concept vectors (\textit{NCAVs}) with the \textit{Concept Miner}. Subsequently, during the testing phase (Fig.~\ref{fig:method-unsupervised}b), the \textit{NCAVs} are utilized to generate \textit{Concept Masks} for activations (\textit{Test Acts}) of \textit{Test Images}.
To evaluate concept similarity via masks, we process obtained continuous \textit{Concept Masks}. Each of them is normalized between 0 and 1, bilinearly interpolated to the same size (e.g., size of corresponding \textit{Test Images}), and then binarized by thresholding, where the threshold value is a hyperparameter.
After completing the preprocessing step, we calculate the \textit{Unsupervised Concept Similarity} ($UCS_{i,j}$) for any pair of concepts by averaging the pixelwise Jaccard index, also known as Intersection over Union (IoU), of set of binary concept masks obtained for test samples:
\begin{align}
    UCS_{i,j} &= \frac{1}{N} \sum_{k=1}^{N}\text{IoU}(M_i^k, M_j^k)
    \;,
    &
    \text{IoU}(M_i^k, M_j^k) &= \frac{\sum \texttt{AND}(M_i^k, M_j^k) }{\sum \texttt{OR}(M_i^k, M_j^k)}
    \;,
\end{align}
where, $i$ and $j$ are concept indices, $N$ is the number of test samples, and $M_i^k, M_j^k\in \{ \text{\texttt{True}}, \text{\texttt{False}} \}^{W\times H}$ are binary concept masks binary interpolated to the same fixed size $W\times H$ (defined by user) for the test sample at index $k$, \texttt{AND} and \texttt{OR} refer to pixel-wise intersection and union of binary masks, respectively.

Therefore, by comparing the projections of extracted concepts onto the input space, we indirectly measure the similarity between concepts and even describe the similarity of latent spaces. By using different test sets to excite and extract desired concepts in various layers, human experts can gain valuable insights into the knowledge similitude across different models.

\input{graphics/main_diagram_supervised}

\subsection{Supervised Feature Space Similarity}
\label{sec:method-supervised-sim}

The supervised approach aims to answer the question, \textit{\enquote{How similar is the arrangement of feature spaces in compared layers with respect to a given concepts?}}. In order to answer it, CAVs~\cite{kim2018interpretability} are utilized as pivot vectors, around which we estimate the behaviour of feature spaces with activations of test samples.

In Figure~\ref{fig:method-supervised}a, the supervised concept-based layer-wise feature space comparison process is shown for two trained \textit{Test Networks}. \textit{L1} and \textit{L2} are indices of analyzed layers. In the first stage, the \textit{Concept Extractor} is employed to extract \textit{CAVs} for each pair of the compared layers, using the training sample activations (\textit{Train Acts}) obtained from concept-related images (\textit{Train Concepts}).
Next (Fig.~\ref{fig:method-supervised}b), to compare the feature spaces with respect to selected concepts, we compute the cosine similarity between the \textit{CAVs} and the activations of test samples (\textit{Test Acts}). Finally, we use the Pearson Correlation Coefficient (PCC) to compare resulting series of cosine similarities and estimate the \textit{Supervised Feature Space Similarity} ($SFSS_{u,v}$), which takes into account the ranking information of the samples as well as accounts for the relative orientation of sample activations in the feature space:
\begin{align}
\label{eq:method-supervised-sim}       
    SFSS_{u,v} &= {
        \frac{1}{M} \sum_{i=1}^{M} }
        \text{PCC}\left(\left\{ \text{CS}_{u,k}^{i} \right\}_{k=1}^N), \left\{ \text{CS}_{v,k}^{i} \right\}_{k=1}^N\right)
    \;
    \\
\label{eq:method-cos-sim}       
    \text{CS}_{*,k}^{i} &= \cos(CAV_*^{i}, x_{*,k})
    \;,~*\in\{u,v\}
\end{align}
where, indices $u$ and $v$ represent network layers, $M$ is the total number of test concepts, $i$ is the index of the currently tested concept, $N$ is the total number of test samples, and $\text{CS}_{*,k}$ is a series of cosine similarities between the tested concept's $CAV$ and the activation $x_{*,k}$ of the $k$-th test sample in layer $*$.

Although we propose using PCC for the computation of $SFSS_{u,v}$, it can be replaced with a statistical metric that preserves the rank order of values in the series. Spearman's rank correlation coefficient, for example, is a valid alternative.

Hence, by ranking and comparing the similarities between concepts representations and test sample activations across multiple layers and models, we can indirectly estimate the generalized similarity and arrangement of the feature spaces in them.

%% file: graphics/main_diagram_unsupervised.tex
\begin{figure*}[t]
  \centering
  \includegraphics[width=\linewidth]{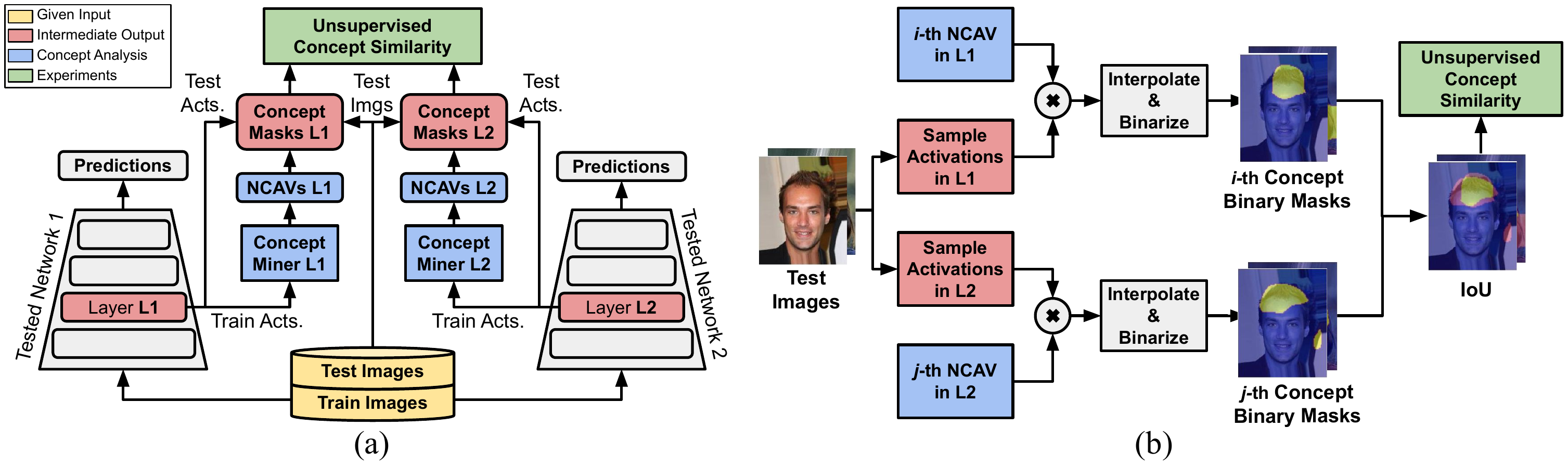}
  \caption{Unsupervised evaluation of concept similarity (a) and concept similarity scoring (b).}
  \label{fig:method-unsupervised}
\end{figure*}

%% file: graphics/main_diagram_supervised.tex
\begin{figure*}[t]
  \centering
  \includegraphics[width=\linewidth]{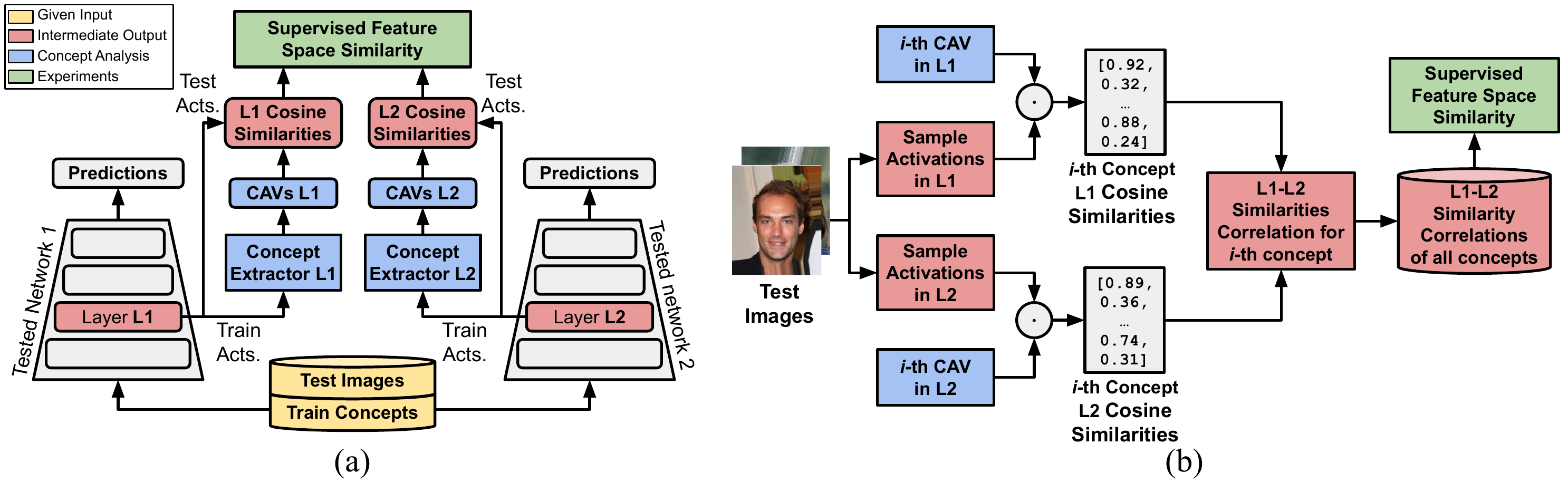}
  \caption{Supervised concept-based estimation of feature space similarity (a) and  feature space similarity scoring (b).}
  \label{fig:method-supervised}
\end{figure*}

%% file: sections/50_setup.tex
\section{Experimental Setup}
\label{sec:setup}

Our experiments follow the methodology outlined in the previous section, which involves two main parts: 1) unsupervised layer-wise estimation of semantic similarity with binary concept masks (Sec.~\ref{sec:method-unsupervised-sim}); and 2) supervised layer-wise comparison of model feature spaces with sample semantic similarity rankings (Sec.~\ref{sec:method-supervised-sim}). In the subsequent subsections, we provide all details on the experimental setup.

\subsection{Experimental Data of Test Images}
\label{sec:setup-datasets}

We assume that the semantic complexity of the test data may affect the performance of the proposed methods. To investigate this, we conduct the evaluation using two datasets with similar knowledge categories but varying semantic complexity: MS COCO 2017~\cite{lin2014microsoft} and CelebA~\cite{liu2015faceattributes}. The CelebA is a low semantic diversity dataset, which comprises over 202,599 homogenous images with celebrity faces. In contrast, the MS COCO dataset is an object detection dataset with high semantic diversity, featuring images of various objects in different contexts. This dataset includes images of different shapes with 2D object bounding box annotations. We utilized a subset of more than 2,000 MS randomly selected COCO images, containing \textit{person} class objects in various positions and situations. To streamline further visual validation, we only used non-crowd instances with bounding box areas of at least $20,000$ pixels. The resulting subset includes more than 2679 bounding boxes of people in different poses and locations extracted from 1685 images.

\subsection{Models}
\label{sec:setup-models}

We perform a semantic comparison of three object detectors of different paradigms and generations, which also feature different backbones, to evaluate the applicability of our approach:
    \begin{itemize}[nosep, leftmargin=1em]
        \item one-stage YOLOv5s\footnote[1]{\url{https://github.com/ultralytics/yolov5}}~\cite{glennjocher20204154370} with residual (res.) DarkNet~\cite{redmon2018yolov3,he2016deep} backbone;
        \item one-stage SSD\footnote[2]{\url{https://pytorch.org/vision/stable/models/ssd}}~\cite{liu2016ssd}, which utilizes a VGG~\cite{simonyan2014very} backbone;
        \item two-stage FasterRCNN\footnote[3]{\url{https://pytorch.org/vision/stable/models/faster_rcnn}}~\cite{ren2015faster} with inverted res. MobileNetV3~\cite{howard2019searching} backbone.
    \end{itemize}
All models are trained on the semantically rich MS COCO~\cite{lin2014microsoft}, which is expected to contain semantic concepts relevant to both test datasets (Sec.~\ref{sec:setup-datasets}).
The models above are further referred to as YOLO5, SSD, and RCNN.

\subsection{Concept Mining and Synthetic Concept Generation}
\label{sec:setup-concepts}

\input{graphics/synthetic_samples}

The effectiveness of supervised concept-based analysis
heavily relies on the quality of the concept-related training data. Unfortunately, publicly available datasets with concept labels are scarce, and existing ones may not be suitable for all research domains and tasks
To address this issue, we suggest generating synthetic concept samples using concept information automatically extracted from task-specific datasets.

For this, we mine concept-related superpixels (image patches) with ICE~\cite{zhang2021invertible} from MS COCO bounding boxes of the \textit{person} class with an area of at least $20,000$ pixels (Sec.~\ref{sec:setup-datasets}). For experiments we selected 3 concepts, each comprising 100 superpixels, and semantically corresponding to labels \enquote{legs}, \enquote{head}, and \enquote{torso}. Concepts were extracted from YOLOv5s layers \texttt{8.v3.c}, \texttt{9.v1.c}, and \texttt{10.c} respectively.


In order to create a synthetic concept sample, 1 to 5 concept-related superpixels are randomly selected, and placed on a background of random noise drawn from a uniform distribution.
Figure~\ref{fig:synth-concepts} shows examples of MS COCO synthetic concepts. Additionally, we rescale the superpixels by a factor between 0.9 and 1.1 before placement. The dimensions of the generated samples are set to be $640 \times 480$ pixels.

To conduct experiments on the CelebA dataset, we utilized four concepts extracted from the \texttt{7.conv} layer of YOLO5 (based on the results of Experiment~1, see Sec.~\ref{sec:results-unsupervised-sim}). These concepts correspond to semantic labels \enquote{hairs}, \enquote{upper face}, \enquote{lower face}, and \enquote{neck}. Each concept sample contains one concept superpixel. The CelebA concept sample has the same size as the dataset sample, i.e., $178 \times 218$ pixels. Top row of Figure~\ref{fig:ice-concepts-celeba-and-mscoco} displays examples of concept masks, which are utilized for concept superpixel cropping.

\subsection{Dimensionality of Concept Activation Vectors}
\label{sec:setup-cav-dims}

\input{graphics/cavs_3d1d}

The TCAV~\cite{kim2018interpretability} method employs 3D-CAVs to represent concepts. However, an alternative approach is to use a 1D-representation, as concept information can be encoded in the linear combination of feature space channels \cite{fong2018net2vec,zhang2021invertible}. The use of a 1D-CAV offers several benefits~\cite{mikriukov2023evaluating} over the 3D-CAV: 1) it is more stable and computationally efficient, as it reduces the number of computational parameters, which is particularly important for a layer-to-layer comparison of deep backbones; 2) it is translation invariant since the spatial information of the concept is aggregated and only the presence or absence of the concept affects the channel activation strength.
Given the mentioned advantages, we have opted to utilize 1D-CAVs in our experiments for supervised feature space comparison (Section~\ref{sec:method-supervised-sim}).


Figure \ref{fig:setup-cav-dims} illustrates the process of obtaining 1D- and 3D-CAVs, where \texttt{C}, \texttt{H}, and \texttt{W} represent the \textit{channel}, \textit{height}, and \textit{width} dimensions, respectively. The arrows indicate the concept extraction process (see Sec.~\ref{sec:background}), where all input representations are aggregated across the \textit{height} and \textit{width} dimensions before computing the 1D-CAV.
When dealing with 1D-representations to compute the similarity between the concept and sample
the sample activation
undergoes the same aggregation across the \textit{height} and \textit{width} dimensions.

\subsection{Experiment-specific Settings}
\label{sec:setup-experiments}

\textbf{Experiment 1:} \textit{Unsupervised Concept Similarity}. We carried out experiments on unsupervised concept similarity using datasets of varying semantic diversity to showcase how concept mining influenced by the input data. To train the NCAVs, we used 100 and 300 random samples from CelebA and MS COCO datasets, respectively, as explained in Section~\ref{sec:method-unsupervised-sim} and Section~\ref{sec:setup-datasets}. Afterward, we evaluated the performance using another 100 samples of each dataset to compute our $UCS_{i,j}$ metric from Section~\ref{sec:method-unsupervised-sim}. The experimental results are presented in the form of a heatmap for each pair of layers.

For CelebA, we extracted 5 concepts per layer, while for MS COCO the number of mined concepts is set to 10. We used a value of $BT=0.25$ (see Sec.~\ref{sec:results-unsupervised-sim} for the analysis of impact of $BT$ values) for binarizing concept masks. Examples of resulting masks for different $BT$ values can be seen in Fig.~\ref{fig:ice-thresholds-samples} in Sec.~\ref{sec:results-unsupervised-sim}.

\noindent\textbf{Experiment 2:} \textit{Supervised Feature Space Similarity}. We evaluated the layer-wise feature space similarity of neural networks by conducting tests using CAVs trained on synthetic concepts from MS COCO and CelebA (see Section~\ref{sec:setup-concepts}). To measure the $SFSS_{i,j}$ ranking metric, we used 200 randomly sampled MS COCO images and the results are presented as a heatmap, with each cell representing a layer combination.
To plot the heatmaps, we selected 10 layers uniformly distributed over the backbone depth of the networks under test (see Sec.~\ref{sec:results-supervised-sim}), which are listed in Table~\ref{tab:tcav_layers}.

\input{graphics/tcav_layers_table}

%% file: graphics/synthetic_samples.tex
\begin{figure*}[t]
     \centering
     \begin{subfigure}[b]{0.27\textwidth}
         \centering
         \includegraphics[width=\textwidth]{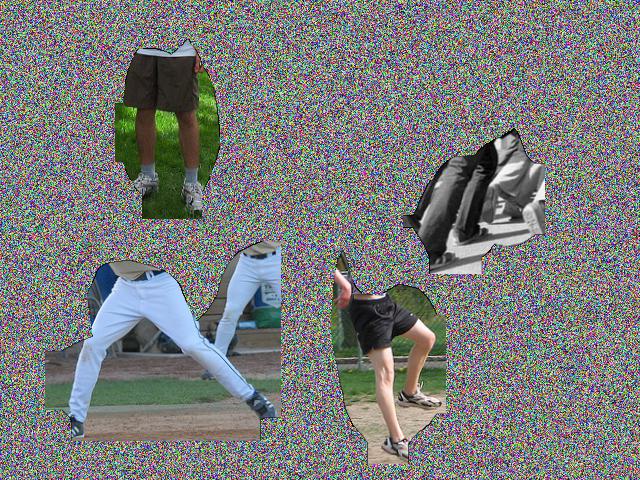}
         \caption{Concept: \enquote{legs}}
         \label{fig:synth-concept-legs}
     \end{subfigure}     
     \hfill     
     \begin{subfigure}[b]{0.27\textwidth}
         \centering
         \includegraphics[width=\textwidth]{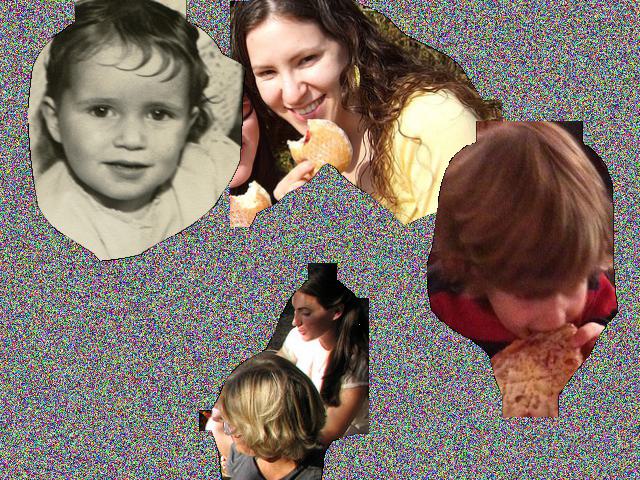}
         \caption{Concept: \enquote{head}}
         \label{fig:synth-concept-head}
     \end{subfigure}     
     \hfill     
     \begin{subfigure}[b]{0.27\textwidth}
         \centering
         \includegraphics[width=\textwidth]{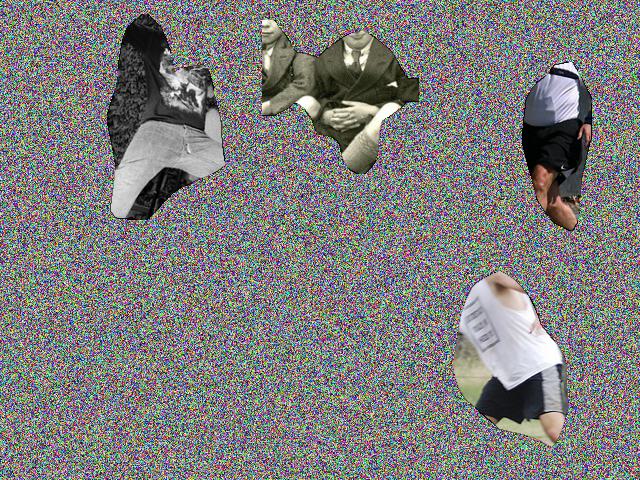}
         \caption{Concept: \enquote{torso}}
         \label{fig:synth-concept-torso}
     \end{subfigure}     
    \caption{Examples of generated MS COCO synthetic concept training samples.}
    \label{fig:synth-concepts}
\end{figure*}

%% file: graphics/cavs_3d1d.tex
%
\begin{figure*}[t]
  \centering
  \includegraphics[width=0.5\linewidth]{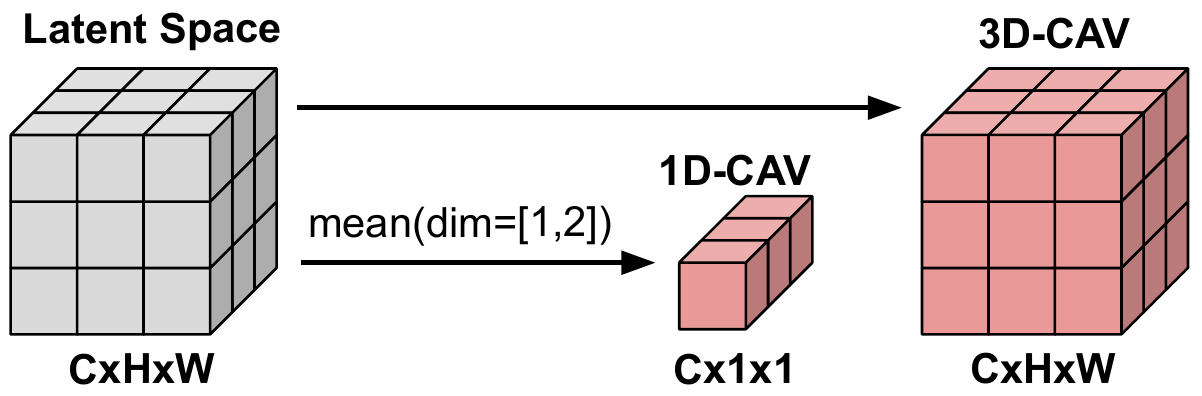}
  \caption{Generation of 3D- and 1D-CAV representations.}
  \label{fig:setup-cav-dims}
\end{figure*}

%% file: graphics/tcav_layers_table.tex
\begin{table}

\setlength{\tabcolsep}{3pt}
\centering
\caption{Shorthands of selected CNN intermediate layers for experiments (b=block, f=features, e=extra, c=conv, v=cv).}
\label{tab:tcav_layers}
\newcommand{\layerID}[1]{\scriptsize{}#1}
\begin{tabular}{c|c|c|c|c|c|c|c|c|c|c|c}
\hline
\multirow{2}{*}{NN} & \multicolumn{10}{c}{Layer id}\\ \cline{2-12}
& $0$ & $1$ & $2$ & $3$ & $4$ & $5$ & $6$ & $7$ & $8$ & $9$ & $10$ \\
\hline
\scriptsize YOLO5
& \layerID{4.v3.c} & \layerID{5.c} & \layerID{6.v3.c} & \layerID{7.c} & \layerID{8.v3.c} & \layerID{9.v2.c} & \layerID{13.v3.c} & \layerID{17.v3.c} & \layerID{18.c} & \layerID{20.v3.c} & \layerID{21.c} \\
\hline
\scriptsize SSD
& \layerID{f.5} & \layerID{f.10} & \layerID{f.14} & \layerID{f.17} & \layerID{f.21} & \layerID{e.0.1} & \layerID{e.0.5} & \layerID{e.1.0} & \layerID{e.2.0} & \layerID{e.3.0} & \layerID{e.4.0} \\
\hline
\scriptsize RCNN
& \layerID{5.b.3} & \layerID{7.b.2} & \layerID{8.b.2} & \layerID{9.b.2} & \layerID{10.b.2} & \layerID{11.b.3} & \layerID{12.b.3} & \layerID{13.b.3} & \layerID{14.b.3} & \layerID{15.b.3} & \layerID{16} \\
\hline

\end{tabular}
\end{table}

%% file: sections/60_experiments.tex
\section{Experimental Results}
\label{sec:results}


\subsection{Unsupervised Concept Similarity}
\label{sec:results-unsupervised-sim}

\input{graphics/ice_samples_celeba_and_mscoco}

The experiments were carried out following the methodology outlined in Section~\ref{sec:method-unsupervised-sim}, using the setup described in Section~\ref{sec:setup-experiments}.
Figures~\ref{fig:ice-comparison},~\ref{fig:ice-comparison-other},~\ref{fig:ice-thresholds-50}, and~\ref{fig:ice-thresholds-75} depict the concept similarity heatmaps of concepts mined in different layers, while Figures~\ref{fig:ice-concepts-celeba-and-mscoco}~and~\ref{fig:ice-thresholds-samples} show examples of the produced concept masks.

\input{graphics/ice_comparison}

\textbf{Impact of data semantic diversity.}
After manually inspecting concept masks and $UCS$-heatmaps generated with CelebA for different layers of the tested CNN backbones, we discovered that YOLO5 \texttt{7.c}, SSD \texttt{e.0.3}, and RCNN \texttt{15.b.3.0} layers (see Tab.\ref{tab:tcav_layers} for shorthands) had the most similar concepts and feature spaces. These are shown in the top row of Fig.~\ref{fig:ice-comparison}. Furthermore, we found a one-to-one correspondence between extracted concepts in these layers. For convenience, we arranged the heatmap's horizontal axis by placing the most similar concept pairs on the diagonal.
All of the extracted concepts are interpretable and correspond to semantic labels  (with optimal layers for them in brackets): 
\enquote{hair} (\texttt{YOLO5.c4}, \texttt{SSD.c4}, \texttt{RCNN.c2}), 
\enquote{upper face} (\texttt{YOLO5.c1}, \texttt{SSD.c2}, \texttt{RCNN.c0}), 
\enquote{lower face} (\texttt{YOLO5.c2}, \texttt{SSD.c0}, \texttt{RCNN.c4}), 
\enquote{neck} (\texttt{YOLO5.c3}, \texttt{SSD.c3}, \texttt{RCNN.c1}), and \enquote{background} (\texttt{YOLO5.c0}, \texttt{SSD.c1}, \texttt{RCNN.c3}).
Figure~\ref{fig:ice-concepts-celeba-and-mscoco}
demonstrates examples of binary masks generated for \enquote{hair}, \enquote{lower face} and \enquote{neck} concepts.

In contrast to CelebA, not all of the concepts mined from MS COCO are human-interpretable and have counterparts in other models. This is demonstrated in the example of layers \texttt{8.v3.c}, \texttt{e.0.5}, and \texttt{15.b.3.0} of YOLO5, SSD, and RCNN (bottom row of Fig.~\ref{fig:ice-comparison}). The higher semantic variability of input samples in MS COCO, where input samples may contain different sets of concepts, makes it more challenging to mine meaningful concepts. However, after a manual inspection of the most similar concept pairs highlighted by our approach, we found concepts corresponding to semantic labels such as \enquote{legs} (\texttt{YOLO5.cc}, \texttt{SSD.c0}, \texttt{RCNN.c2}), \enquote{head} (\texttt{YOLO5.c2}, \texttt{SSD.c1}, \texttt{RCNN.c1}), and \enquote{background} (\texttt{YOLO5.c8}, \texttt{SSD.c8}, \texttt{RCNN.c7}). Examples of their binary masks are depicted in
the bottom row of Figure~\ref{fig:ice-concepts-celeba-and-mscoco}.

To summarize our observations, we conclude that datasets with high semantic variability may lead to a lower quality of automatic concept extraction results. Our proposed method can quantify and visualize this issue, assist in finding similar concepts, and identify layers with similar semantic information. Also, for a comprehensive unsupervised comparison of model concepts and feature spaces, we recommend using datasets with semantically homogeneous samples, similar to those found in CelebA.

\input{graphics/ice_comparison_other}

\textbf{Semantically similar layers identification.}
The proposed method enables the identification of the most similar layers in different networks. For example, the top row diagrams of Figure~\ref{fig:ice-comparison} display the layers with the highest level of feature space correspondence for CelebA, where each concept of one layer has a distinct counterpart in another. Another example in Figure~\ref{fig:ice-comparison-other} demonstrates non-optimal variations where a concept has multiple possible counterparts (Fig.~\ref{fig:ice-celeba-yolo-rcnn-other}) or no matches (Fig.~\ref{fig:ice-celeba-yolo-ssd-other}).

In our experience, the main factors that influence the identification of similar layers are the number of concepts mined and the semantic complexity of the test dataset, as demonstrated in Figure~\ref{fig:ice-comparison}.

\input{graphics/ice_concept_thresholds}

\textbf{Concept robustness with regards to $BT$.}
The parameters of binary masks generated by ICE on test samples depend on the binarization threshold $BT$: higher $BT$ values may reduce mask size and, hence, impact concept similarity, as illustrated in Fig.~\ref{fig:ice-thresholds-samples}.
By leveraging this finding, we can also quantify the relative robustness of different concepts. As illustrated in Figures \ref{fig:ice-thresholds-50} and \ref{fig:ice-thresholds-75}, concepts like \texttt{YOLO5.c1} and \texttt{SSD.c2}, as well as \texttt{YOLO5.c2} and \texttt{SSD.c0}, exhibit the most resilience to changes in $BT$, making them the most robust ones.

\input{graphics/tcav_comparison_all_layers}
\input{graphics/tcav_comparison}

\subsection{Supervised Feature Space Similarity}
\label{sec:results-supervised-sim}

\textbf{Semantic information flow.}
Figures~\ref{fig:tcav-comparison-all-layers}~and~\ref{fig:tcav-comparison} display the layer-wise similarity between the feature spaces of models with respect to given concepts. Notably, the diagonal values in the heatmap of Figure~\ref{fig:tcav-comparison-all-layers} are more intense, indicating that semantic similarity is primarily influenced by the layer's relative depth in the backbone. Therefore, we can compare entire networks by evaluating a selected set of N layers (like in Table~\ref{tab:tcav_layers}) that are evenly distributed throughout the backbone. Such approach helps save processing power and time while preserving the global picture.

\textbf{Concept complexity.} By examining Figure~\ref{fig:tcav-comparison}, we can observe that concepts derived from the CelebA dataset, which are lower in abstraction level and pertain to different parts of the face, lead to greater similarity between layers in identical model pairs compared to the more complex body part concepts extracted from MS COCO. Additionally, these concepts are more clearly defined across a broader range of layers, resulting in distinguishable clusters (larger darker regions) on the heatmaps. These observations imply that these concepts are more effectively represented in the feature spaces of the compared models.

\textbf{Network architecture differences.}
Among the tested backbones, the MobileNetV3 backbone of RCNN exhibits a remarkably distinct behavior. Specifically, MobileNetV3 captures the same semantic information in two distinct regions within the network: at the beginning and in the middle. This can be observed in the middle and right columns of Figure~\ref{fig:tcav-comparison}, where we see a pattern with two distinct clusters (darker areas) along the vertical axis, between layers 0 and 3, and layers 5 and 8. This pattern is not observed in the direct comparison of the DarkNet and VGG backbones of YOLO5 and SSD, and, thus, only typical to MobileNetV3. We attribute this peculiarity to the distinctive network-building technique employed in inverted residual blocks of MobileNetV3, which allows to propagate semantic information of tested concepts across the network more effectively. This, in turn, leads to a comparatively lesser decrease in semantic similarity of deeper layers in RCNN than in SSD and YOLO5.

Thereby, the proposed method for supervised feature space comparison also allows us to identify significant variations in the feature spaces and semantic representation learning across different models, and also can be used to judge on optimal model architectures with respect to interpretability.

\subsection{Limitations and Future Work}
\label{sec:results-limitations}

Our approaches for concept analysis naturally inherit all limitations of data-driven methods, like dependence on high-quality data. Thus, manual visual validation of used CAVs and NCAVs, as done here, remains inevitable. Our proposed semi-automatic data generation can be used to reduce labeling costs.
Moreover, we found that the semantic diversity of the test data strongly affects the quality of the extracted concepts, and hence recommend using semantically homogeneous sets for testing. 

A limitation inherent to using ICE concept masks is the differing and low mask resolution resulting from the different activation map dimensions. Choosing the scaling factors individually for any pair of layers may mitigate this, however, at cost of comparability.
Another issue is the dependence on the binarization threshold $BT$. An interesting future direction could therefore be to directly compare the non-binary concept masks.

In general, it will be interesting to apply our approach to further large NN architectures, e.g., transformers, and other visual tasks than object detection.

%% file: graphics/ice_samples_celeba_and_mscoco.tex
\begin{figure*}[t]
  \centering
  \includegraphics[width=0.9\linewidth]{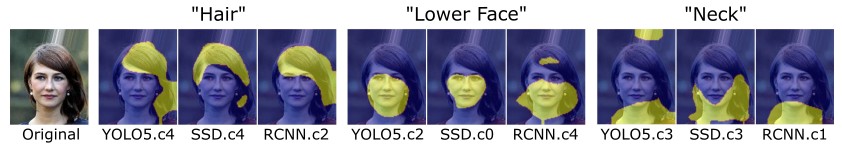}\\%
  \includegraphics[width=0.9\linewidth]{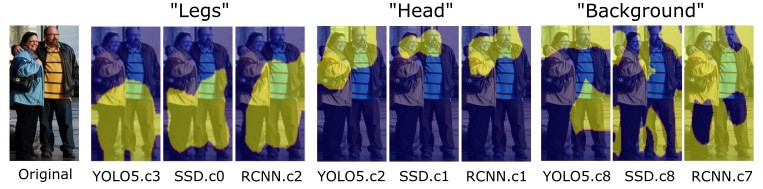}
  \caption{Examples of binary concept masks obtained unsupervised using ICE for CelebA (top) and MS COCO (bottom) at binarization threshold $BT=0.25$.}
  \label{fig:ice-concepts-celeba-and-mscoco}
\end{figure*}

%% file: graphics/ice_comparison.tex

\begin{figure*}[t]
  \centering
  \includegraphics[width=\linewidth]{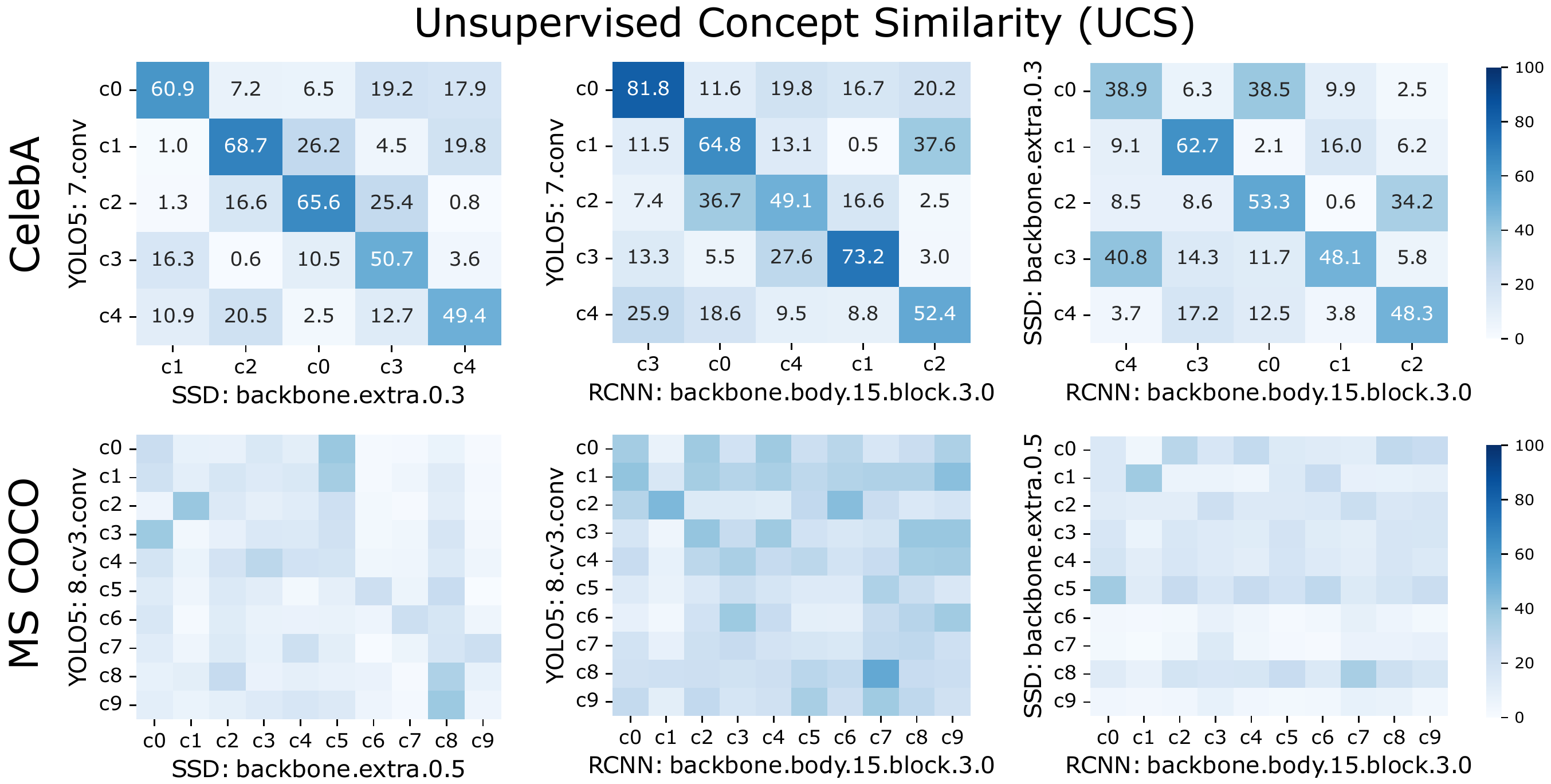}
  \caption{Unsupervised concept similarity ($UCS_{i,j}$) estimates of different concepts $c_i$ (x-axis) and $c_j$ (y-axis) mined in \textit{optimal layers}.}
  \label{fig:ice-comparison}
\end{figure*}

%% file: graphics/ice_comparison_other.tex

\begin{figure*}[t]
     \centering
     \begin{subfigure}[b]{0.38\textwidth}
         \centering
         \includegraphics[width=\textwidth]{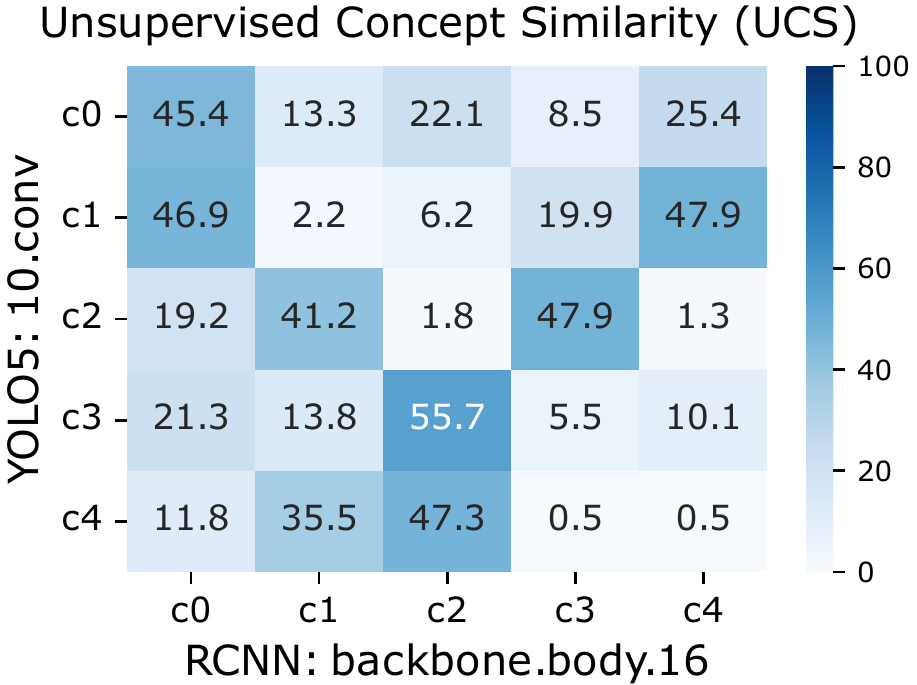}
         \caption{Several concept counterparts.}
         \label{fig:ice-celeba-yolo-rcnn-other}
     \end{subfigure}
     \hspace{5mm}
     \begin{subfigure}[b]{0.38\textwidth}
         \centering
         \includegraphics[width=\textwidth]{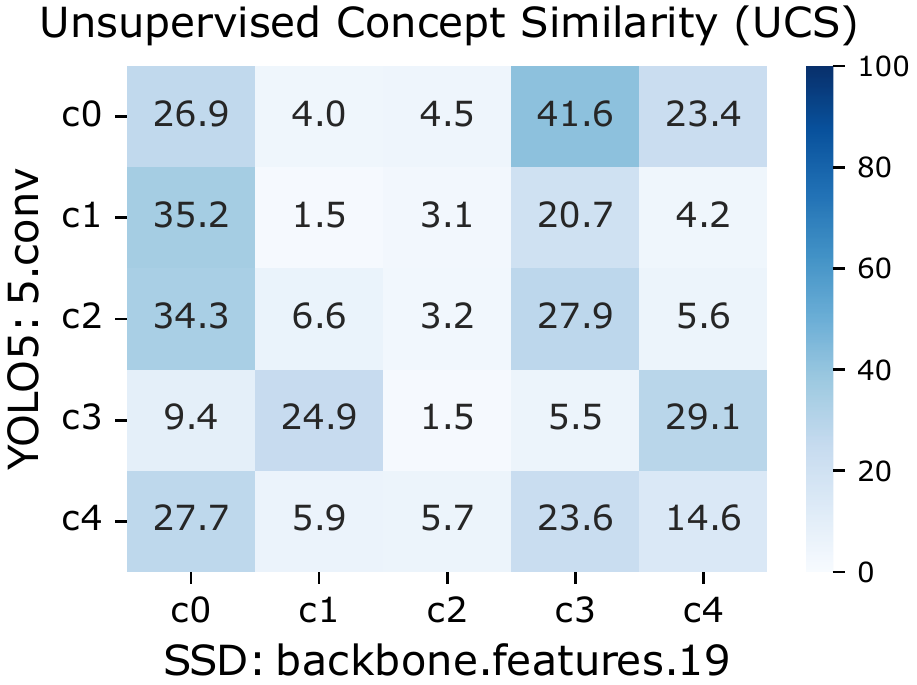}
         \caption{\texttt{SSD.c2} has no counterpart.}
         \label{fig:ice-celeba-yolo-ssd-other}
    \end{subfigure}
    \caption{Unsupervised concept similarity ($UCS_{i,j}$) estimates of different concepts $c_i$ (x-axis) and $c_j$ (y-axis) mined in \textit{non-optimal layers}.}
    \label{fig:ice-comparison-other}
\end{figure*}

%% file: graphics/ice_concept_thresholds.tex
\begin{figure*}[t]
     \centering
     \begin{subfigure}[b]{0.32\textwidth}
         \centering
         \includegraphics[width=\textwidth]{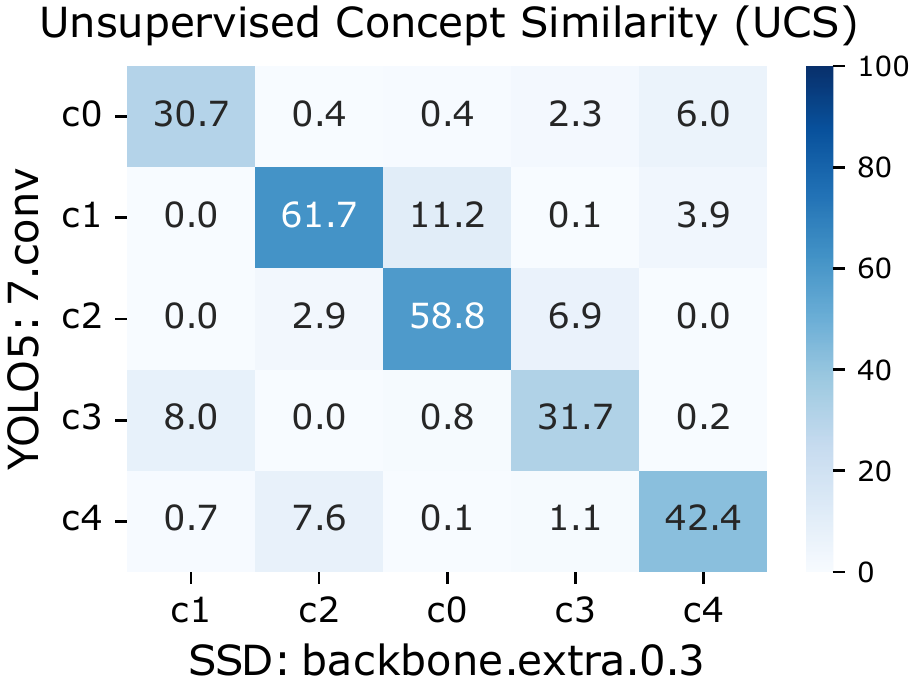}
         \caption{$BT=0.5$}
         \label{fig:ice-thresholds-50}
     \end{subfigure}     
     \hfill     
     \begin{subfigure}[b]{0.32\textwidth}
         \centering
         \includegraphics[width=\textwidth]{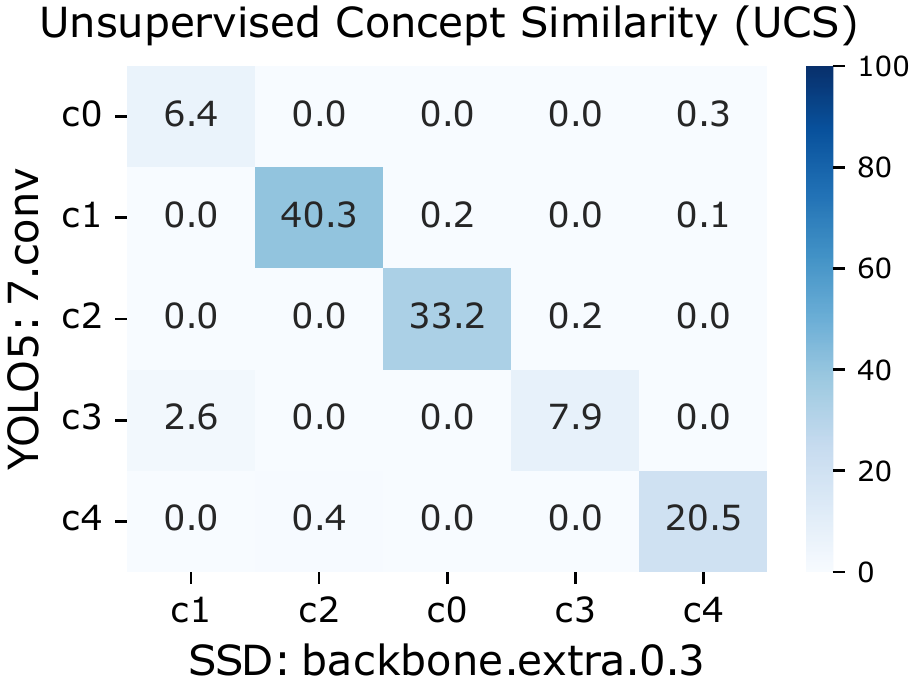}
         \caption{$BT=0.75$}
         \label{fig:ice-thresholds-75}
     \end{subfigure}     
     \hfill     
     \begin{subfigure}[b]{0.31\textwidth}
         \centering
         \includegraphics[width=\textwidth]{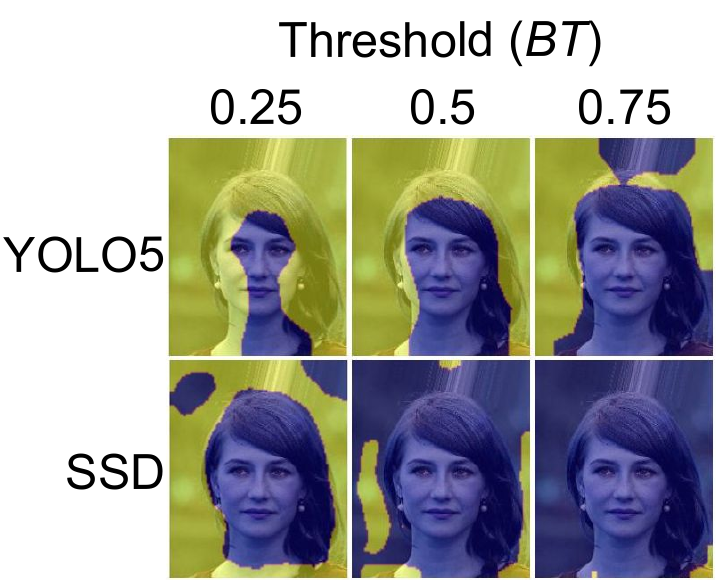}
         \caption{\texttt{YOLO5.c0} \& \texttt{SSD.c1}}
         \label{fig:ice-thresholds-samples}
     \end{subfigure}     
    \caption{Influence of concept mask binarization threshold value ($BT$) on unsupervised concept similarity estimation of concepts $c_i$ (x-axis) and $c_j$ (y-axis) for \texttt{7.conv} and \texttt{backbone.extra.0.3} layers of YOLO5 and SSD.}
    \label{fig:ice-thresholds}
\end{figure*}

%% file: graphics/tcav_comparison_all_layers.tex

\begin{figure*}[t]     
    \centering
    \includegraphics[width=\textwidth]{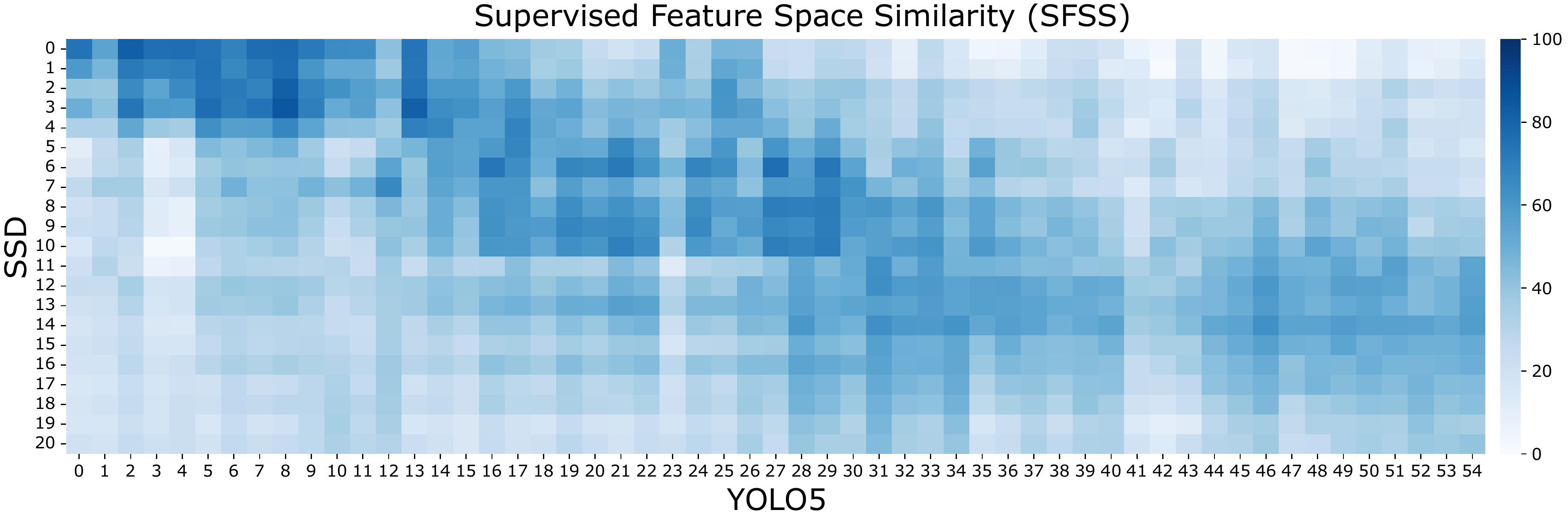}
    \caption{Supervised feature space comparison of SSD and YOLO5 layers (all convolutional layers indexed from 0 to 20 resp.\ 54).}
    \label{fig:tcav-comparison-all-layers}
\end{figure*}

%% file: graphics/tcav_comparison.tex

\begin{figure*}[t]
  \centering
  \includegraphics[width=\linewidth]{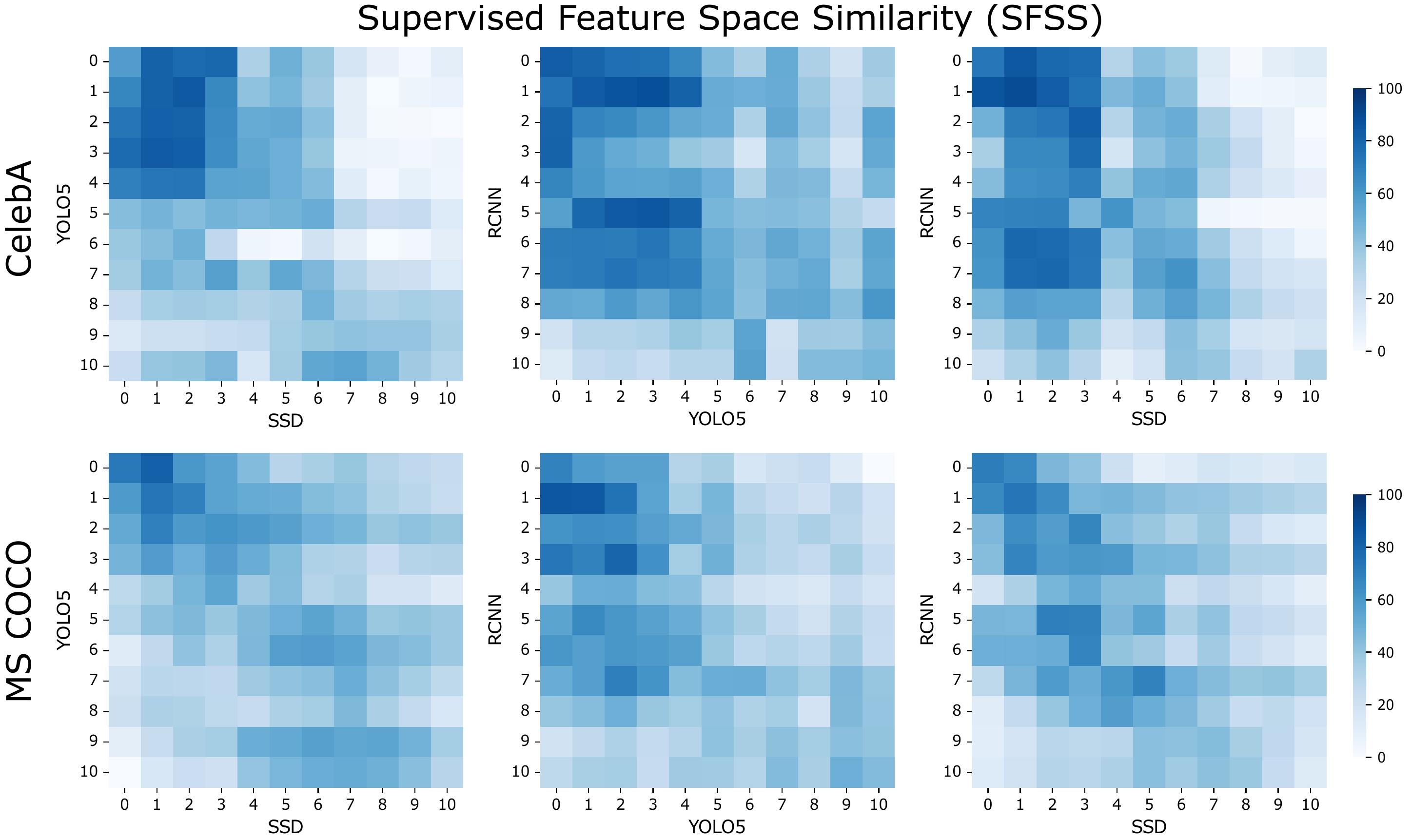}
  \caption{Supervised feature space comparison of selected model layers (cf. Tab.~\ref{tab:tcav_layers}).}
  \label{fig:tcav-comparison}
\end{figure*}

%% file: sections/70_conclusion.tex
\section{Conclusion and Outlook}
\label{sec:conclusion}

In this research, we presented architecture-agnostic supervised and unsupervised methods for estimating the similarity of feature spaces in CNN backbones. Proposed methods help to reveal how the same semantic information is processed across various model backbones, and enable identification of the semantically similar layers. We use semantic concept vectors, namely CAVs and NCAVs, to assess the behavior of the latent space through the concept's response to the test data. Experiments on two datasets and three different backbone architectures trained on the same data
revealed that regardless of the NN architecture, layers with similar semantic information can be found, as we found network layers with one-to-one concept correspondence. We also discovered that the feature space semantic information depends on the relative depth of the layer in the network backbone. Therefore, to compare different CNN backbones, it seems to be sufficient to compare only a subset of layers of uniform depth-distance in the backbone. Finally, our method provides valuable insights, which may be useful for applications like informed model selection, meta-analysis of network architectures, or dataset bias retrieval.

%% file: sections/80_acknowledgments.tex
\subsection*{Acknowledgments}

The research leading to these results is funded by the German Federal Ministry for Economic Affairs and Climate Action within the project “KI Wissen – Entwicklung von Methoden für die Einbindung von Wissen in maschinelles Lernen”. The authors would like to thank the consortium for the successful cooperation.

%% file: main.bbl
\begin{thebibliography}{10}
\providecommand{\url}[1]{\texttt{#1}}
\providecommand{\urlprefix}{URL }
\providecommand{\doi}[1]{https://doi.org/#1}

\bibitem{iso26262}
32, I.S.: {ISO 26262-1:2018(En): Road Vehicles -- Functional Safety -- Part 1:
  Vocabulary} (2018), \url{https://www.iso.org/standard/68383.html}

\bibitem{achtibat2022towards}
Achtibat, R., Dreyer, M., Eisenbraun, I., Bosse, S., Wiegand, T., Samek, W.,
  Lapuschkin, S.: From" where" to" what": Towards human-understandable
  explanations through concept relevance propagation. arXiv preprint
  arXiv:2206.03208  (2022)

\bibitem{arrieta2020explainable}
Arrieta, A.B., D{\'\i}az-Rodr{\'\i}guez, N., Del~Ser, J., Bennetot, A., Tabik,
  S., Barbado, A., Garc{\'\i}a, S., Gil-L{\'o}pez, S., Molina, D., Benjamins,
  R., et~al.: Explainable artificial intelligence (xai): Concepts, taxonomies,
  opportunities and challenges toward responsible ai. Information fusion
  \textbf{58},  82--115 (2020)

\bibitem{bach2015pixel}
Bach, S., Binder, A., Montavon, G., Klauschen, F., M{\"u}ller, K.R., Samek, W.:
  On pixel-wise explanations for non-linear classifier decisions by layer-wise
  relevance propagation. PloS one  \textbf{10}(7),  e0130140 (2015)

\bibitem{bau2017network}
Bau, D., Zhou, B., Khosla, A., Oliva, A., Torralba, A.: Network dissection:
  Quantifying interpretability of deep visual representations. In: Proc. IEEE
  conf. computer vision and pattern recognition. pp. 6541--6549 (2017)

\bibitem{bolya2020tide}
Bolya, D., Foley, S., Hays, J., Hoffman, J.: Tide: A general toolbox for
  identifying object detection errors. In: Computer Vision--ECCV 2020: 16th
  European Conference, Glasgow, UK, August 23--28, 2020, Proceedings, Part III
  16. pp. 558--573. Springer (2020)

\bibitem{chen2019looks}
Chen, C., Li, O., Tao, D., Barnett, A., Rudin, C., Su, J.K.: This looks like
  that: deep learning for interpretable image recognition. Advances in neural
  information processing systems  \textbf{32} (2019)

\bibitem{chyung_extracting_2019}
Chyung, C., Tsang, M., Liu, Y.: Extracting interpretable concept-based decision
  trees from {{CNNs}}. In: Proc. 2019 {{ICML Workshop Human}} in the {{Loop
  Learning}}. vol. 1906.04664. {CoRR} (Jun 2019)

\bibitem{esser_disentangling_2020}
Esser, P., Rombach, R., Ommer, B.: A disentangling invertible interpretation
  network for explaining latent representations. In: Proc. 2020 {{IEEE Conf}}.
  {{Comput}}. {{Vision}} and {{Pattern Recognition}}. pp. 9220--9229. {IEEE}
  (Jun 2020). \doi{10.1109/CVPR42600.2020.00924}

\bibitem{fong2018net2vec}
Fong, R., Vedaldi, A.: Net2vec: Quantifying and explaining how concepts are
  encoded by filters in deep neural networks. In: Proc. IEEE conf. computer
  vision and pattern recognition. pp. 8730--8738 (2018)

\bibitem{ge2021peek}
Ge, Y., Xiao, Y., Xu, Z., Zheng, M., Karanam, S., Chen, T., Itti, L., Wu, Z.: A
  peek into the reasoning of neural networks: Interpreting with structural
  visual concepts. In: Proceedings of the IEEE/CVF Conference on Computer
  Vision and Pattern Recognition. pp. 2195--2204 (2021)

\bibitem{ghorbani2019towards}
Ghorbani, A., Wexler, J., Zou, J.Y., Kim, B.: Towards automatic concept-based
  explanations. Advances in Neural Information Processing Systems  \textbf{32}
  (2019)

\bibitem{giunchiglia_roadr_2022}
Giunchiglia, E., Stoian, M., Khan, S., Cuzzolin, F., Lukasiewicz, T.:
  {{ROAD-R}}: {{The Autonomous Driving Dataset}} with {{Logical Requirements}}.
  In: {{IJCLR}} 2022 {{Workshops}} (Jun 2022)

\bibitem{goodman2017european}
Goodman, B., Flaxman, S.: European union regulations on algorithmic
  decision-making and a ``right to explanation''. AI Magazine  \textbf{38}(3),
  50--57 (Oct 2017). \doi{10.1609/aimag.v38i3.2741},
  \url{https://ojs.aaai.org/index.php/aimagazine/article/view/2741}

\bibitem{graziani_concept_2020}
Graziani, M., Andrearczyk, V., {Marchand-Maillet}, S., M{\"u}ller, H.: Concept
  attribution: {{Explaining CNN}} decisions to physicians. Computers in Biology
  and Medicine  \textbf{123},  103865 (Aug 2020).
  \doi{10.1016/j.compbiomed.2020.103865}

\bibitem{gu_semantics_2019}
Gu, J., Tresp, V.: Semantics for global and local interpretation of deep neural
  networks. CoRR  \textbf{abs/1910.09085} (Oct 2019)

\bibitem{he2016deep}
He, K., Zhang, X., Ren, S., Sun, J.: Deep residual learning for image
  recognition. In: Proc. IEEE onf. computer vision and pattern recognition. pp.
  770--778 (2016)

\bibitem{hohman_summit_2020}
Hohman, F., Park, H., Robinson, C., Polo~Chau, D.H.: Summit: {{Scaling Deep
  Learning Interpretability}} by {{Visualizing Activation}} and {{Attribution
  Summarizations}}. IEEE Transactions on Visualization and Computer Graphics
  \textbf{26}(1),  1096--1106 (Jan 2020). \doi{10.1109/TVCG.2019.2934659}

\bibitem{howard2019searching}
Howard, A., Sandler, M., Chu, G., Chen, L.C., Chen, B., Tan, M., Wang, W., Zhu,
  Y., Pang, R., Vasudevan, V., et~al.: Searching for mobilenetv3. In:
  Proceedings of the IEEE/CVF international conference on computer vision. pp.
  1314--1324 (2019)

\bibitem{glennjocher20204154370}
Jocher, G.: {YOLOv5 in PyTorch, ONNX, CoreML, TFLite}.
  \url{https://github.com/ultralytics/yolov5} (Oct 2020).
  \doi{10.5281/zenodo.4154370}, \url{https://doi.org/10.5281/zenodo.4154370}

\bibitem{kazhdan_now_2020}
Kazhdan, D., Dimanov, B., Jamnik, M., Li{\`o}, P., Weller, A.: Now you see me
  ({{CME}}): {{Concept-based}} model extraction. In: Proc. 29th {{ACM Int}}.
  {{Conf}}. {{Information}} and {{Knowledge Management Workshops}}. {{CEUR}}
  Workshop Proceedings, vol.~2699. {CEUR-WS.org} (2020)

\bibitem{kim2018interpretability}
Kim, B., Wattenberg, M., Gilmer, J., Cai, C., Wexler, J., Viegas, F., et~al.:
  Interpretability beyond feature attribution: Quantitative testing with
  concept activation vectors (tcav). In: Int. conf. machine learning. pp.
  2668--2677. PMLR (2018)

\bibitem{koh2020concept}
Koh, P.W., Nguyen, T., Tang, Y.S., Mussmann, S., Pierson, E., Kim, B., Liang,
  P.: Concept bottleneck models. In: Int. conf. Machine Learning. pp.
  5338--5348. PMLR (2020)

\bibitem{lapuschkin_unmasking_2019}
Lapuschkin, S., W{\"a}ldchen, S., Binder, A., Montavon, G., Samek, W.,
  M{\"u}ller, K.R.: Unmasking {{Clever Hans}} predictors and assessing what
  machines really learn. Nature Communications  \textbf{10}(1), ~1096 (Mar
  2019). \doi{10.1038/s41467-019-08987-4}

\bibitem{lin2014microsoft}
Lin, T.Y., Maire, M., Belongie, S., Hays, J., Perona, P., Ramanan, D.,
  Doll{\'a}r, P., Zitnick, C.L.: Microsoft coco: Common objects in context. In:
  European onf. computer vision. pp. 740--755. Springer (2014)

\bibitem{liu2016ssd}
Liu, W., Anguelov, D., Erhan, D., Szegedy, C., Reed, S., Fu, C.Y., Berg, A.C.:
  Ssd: Single shot multibox detector. In: Computer Vision--ECCV 2016: 14th
  European Conference, Amsterdam, The Netherlands, October 11--14, 2016,
  Proceedings, Part I 14. pp. 21--37. Springer (2016)

\bibitem{liu2015faceattributes}
Liu, Z., Luo, P., Wang, X., Tang, X.: Deep learning face attributes in the
  wild. In: Proceedings of International Conference on Computer Vision (ICCV)
  (December 2015)

\bibitem{losch_interpretability_2019}
Losch, M., Fritz, M., Schiele, B.: Interpretability beyond classification
  output: {{Semantic}} bottleneck networks. In: Proc. 3rd {{ACM Computer
  Science}} in {{Cars Symp}}. {{Extended Abstracts}} (Oct 2019)

\bibitem{lucieri_explaining_2020}
Lucieri, A., Bajwa, M.N., Dengel, A., Ahmed, S.: Explaining {{AI-based}}
  decision support systems using concept localization maps. In: Neural
  {{Information Processing}}. pp. 185--193. Communications in {{Computer}} and
  {{Information Science}}, {Springer International Publishing} (2020).
  \doi{10.1007/978-3-030-63820-7_21}

\bibitem{mikriukov2023evaluating}
Mikriukov, G., Schwalbe, G., Hellert, C., Bade, K.: Evaluating the stability of
  semantic concept representations in cnns for robust explainability. arXiv
  preprint arXiv:2304.14864  (2023)

\bibitem{miller2022s}
Miller, D., Moghadam, P., Cox, M., Wildie, M., Jurdak, R.: What’s in the
  black box? the false negative mechanisms inside object detectors. IEEE
  Robotics and Automation Letters  \textbf{7}(3),  8510--8517 (2022)

\bibitem{mittelstadt2019explaining}
Mittelstadt, B., Russell, C., Wachter, S.: Explaining explanations in ai. In:
  Proc. conf. fairness, accountability, and transparency. pp. 279--288 (2019)

\bibitem{nguyen_understanding_2019}
Nguyen, A., Yosinski, J., Clune, J.: Understanding {{Neural Networks}} via
  {{Feature Visualization}}: {{A Survey}}. In: Explainable {{AI}}:
  {{Interpreting}}, {{Explaining}} and {{Visualizing Deep Learning}}, pp.
  55--76. Lecture {{Notes}} in {{Computer Science}}, {Springer International
  Publishing} (2019). \doi{10.1007/978-3-030-28954-6_4}

\bibitem{rabold_expressive_2020}
Rabold, J., Schwalbe, G., Schmid, U.: Expressive explanations of dnns by
  combining concept analysis with ilp. In: KI 2020: Advances in Artificial
  Intelligence. pp. 148--162. Lecture Notes in Computer Science, Springer
  International Publishing (2020). \doi{10.1007/978-3-030-58285-2_11},
  \url{https://arxiv.org/abs/2105.07371}

\bibitem{rabold_explaining_2018}
Rabold, J., Siebers, M., Schmid, U.: Explaining black-box classifiers with
  {{ILP}} \textendash{} empowering {{LIME}} with {{Aleph}} to approximate
  non-linear decisions with relational rules. In: Proc. {{Int}}. {{Conf}}.
  {{Inductive Logic Programming}}. pp. 105--117. Lecture {{Notes}} in
  {{Computer Science}}, {Springer International Publishing} (2018).
  \doi{10.1007/978-3-319-99960-9_7}

\bibitem{redmon2018yolov3}
Redmon, J., Farhadi, A.: Yolov3: An incremental improvement. arXiv preprint
  arXiv:1804.02767  (2018)

\bibitem{ren2015faster}
Ren, S., He, K., Girshick, R., Sun, J.: Faster r-cnn: Towards real-time object
  detection with region proposal networks. Advances in neural information
  processing systems  \textbf{28} (2015)

\bibitem{ribeiro2016model}
Ribeiro, M.T., Singh, S., Guestrin, C.: Model-agnostic interpretability of
  machine learning. arXiv preprint arXiv:1606.05386  (2016)

\bibitem{rudin_stop_2019}
Rudin, C.: Stop explaining black box machine learning models for high stakes
  decisions and use interpretable models instead. Nature Machine Intelligence
  \textbf{1}(5),  206--215 (May 2019). \doi{10.1038/s42256-019-0048-x}

\bibitem{schwalbe_concept_2022}
Schwalbe, G.: Concept {{Embedding Analysis}}: {{A Review}}. arXiv:2203.13909
  [cs, stat]  (Mar 2022)

\bibitem{schwalbe2021comprehensive}
Schwalbe, G., Finzel, B.: A comprehensive taxonomy for explainable artificial
  intelligence: A systematic survey of surveys on methods and concepts. arXiv
  e-prints pp. arXiv--2105 (2021)

\bibitem{schwalbe_enabling_2022}
Schwalbe, G., Wirth, C., Schmid, U.: Enabling verification of deep neural
  networks in perception tasks using fuzzy logic and concept embeddings (Mar
  2022)

\bibitem{selvaraju2017grad}
Selvaraju, R.R., Cogswell, M., Das, A., Vedantam, R., Parikh, D., Batra, D.:
  Grad-cam: Visual explanations from deep networks via gradient-based
  localization. In: Proc. IEEE int. conf. computer vision. pp. 618--626 (2017)

\bibitem{simonyan2014very}
Simonyan, K., Zisserman, A.: Very deep convolutional networks for large-scale
  image recognition. arXiv preprint arXiv:1409.1556  (2014)

\bibitem{varghese_unsupervised_2021}
Varghese, S., Gujamagadi, S., Klingner, M., Kapoor, N., B{\"a}r, A., Schneider,
  J.D., Maag, K., Schlicht, P., H{\"u}ger, F., Fingscheidt, T.: An unsupervised
  temporal consistency ({{TC}}) loss to improve the performance of semantic
  segmentation networks. In: 2021 {{IEEE}}/{{CVF Conf}}. {{Comput}}. {{Vision}}
  and {{Pattern Recognition Workshops}}. pp. 12--20 (Jun 2021).
  \doi{10.1109/CVPRW53098.2021.00010}

\bibitem{wan_nbdt_2020}
Wan, A., Dunlap, L., Ho, D., Yin, J., Lee, S., Petryk, S., Bargal, S.A.,
  Gonzalez, J.E.: {{NBDT}}: {{Neural-backed}} decision tree. In: Posters 2021
  {{Int}}. {{Conf}}. {{Learning Representations}} (Sep 2020)

\bibitem{wang2022hint}
Wang, A., Lee, W.N., Qi, X.: Hint: Hierarchical neuron concept explainer. In:
  Proceedings of the IEEE/CVF Conference on Computer Vision and Pattern
  Recognition. pp. 10254--10264 (2022)

\bibitem{wang_chain_2020}
Wang, D., Cui, X., Wang, Z.J.: {{CHAIN}}: {{Concept-harmonized}} hierarchical
  inference interpretation of deep convolutional neural networks. CoRR
  \textbf{abs/2002.01660} (2020)

\bibitem{zhang_interpreting_2018}
Zhang, Q., Cao, R., Shi, F., Wu, Y.N., Zhu, S.C.: Interpreting {{CNN}}
  knowledge via an explanatory graph. In: Proc. 32nd {{AAAI Conf}}.
  {{Artificial Intelligence}}. pp. 4454--4463. {AAAI Press} (2018)

\bibitem{zhang2021invertible}
Zhang, R., Madumal, P., Miller, T., Ehinger, K.A., Rubinstein, B.I.: Invertible
  concept-based explanations for cnn models with non-negative concept
  activation vectors. In: Proc. AAAI Conf. Artificial Intelligence. pp.
  11682--11690 (2021)

\bibitem{zhou2016learning}
Zhou, B., Khosla, A., Lapedriza, A., Oliva, A., Torralba, A.: Learning deep
  features for discriminative localization. In: Proc. IEEE conf. computer
  vision and pattern recognition. pp. 2921--2929 (2016)

\end{thebibliography}
